%% file: main.tex
\documentclass[10pt,twocolumn,letterpaper]{article}

\usepackage[pagenumbers]{cvpr} 


\definecolor{cvprblue}{rgb}{0.21,0.49,0.74}
\usepackage[pagebackref,breaklinks,colorlinks,allcolors=cvprblue]{hyperref}

\def\paperID{10872} 
\def\confName{CVPR}
\def\confYear{2025}

\title{Concept Replacer: Replacing Sensitive Concepts in Diffusion Models via Precision Localization}

\author{Lingyun Zhang\\
Fudan University\\
{\tt\small lyzhang22@m.fudan.edu.cn }
\and
Yu Xie\thanks{*Corresponding authors.} \\
Purple Mountain Laboratories\\
{\tt\small yxie18@fudan.edu.cn }
\and
Yanwei Fu \\
Fudan University\\
{\tt\small yanweifu@fudan.edu.cn}
\and
Ping Chen$^{*}$\\
Fudan University\\
{\tt\small pchen@fudan.edu.cn}
}

\begin{document}

\twocolumn[{%
\renewcommand\twocolumn[1][]{#1}%
\maketitle

\begin{center}
    \centering
    \captionsetup{type=figure}
    \includegraphics[width=0.95 \textwidth]{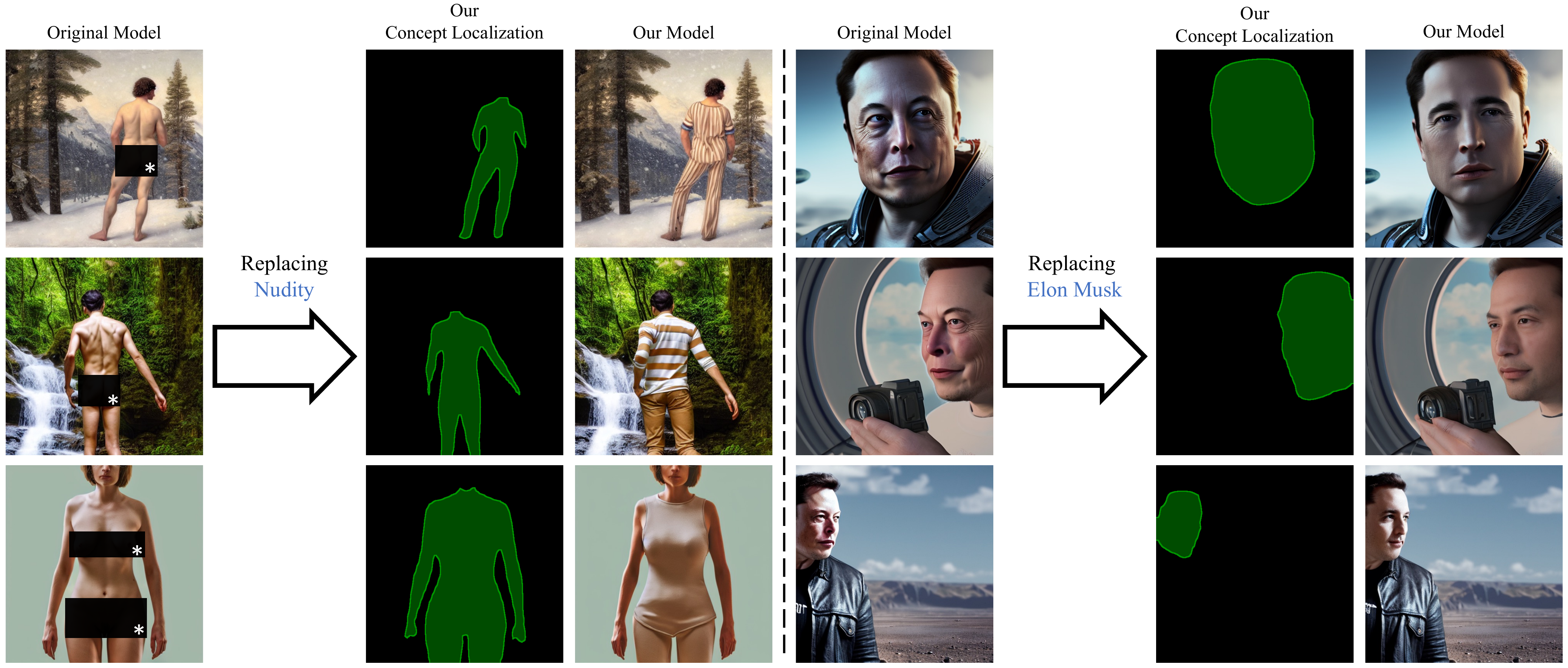}
    \vspace{-0.1in}
    \caption{Given a specified concept for replacement, our method precisely identifies its location during the generation phase and seamlessly replaces it, ensuring that non-target regions remain unaffected. Sensitive areas were masked in black by authors.}
    {\label{teaser}}
\end{center}%
}]


\renewcommand{\thefootnote}{}%
\footnotetext{*Corresponding authors.}%
\renewcommand{\thefootnote}{\arabic{footnote}}

\input{sec/0_abstract}

\input{sec/1_intro}

\input{sec/2_ralated_work}

\input{sec/3_method}

\input{sec/4_experiments}
\input{sec/5_ablation}

\input{sec/6_conclusion}
{
    \small
    \bibliographystyle{ieeenat_fullname}
    \bibliography{main}
}


\end{document}

%% file: sec/0_abstract.tex
\begin{abstract}

As large-scale diffusion models continue to advance, they excel at producing high-quality images but often generate unwanted content, such as sexually explicit or violent content. Existing methods for concept removal generally guide the image generation process but can unintentionally modify unrelated regions, leading to inconsistencies with the original model. We propose a novel approach for targeted concept replacing in diffusion models, enabling specific concepts to be removed without affecting non-target areas. Our method introduces a dedicated concept localizer for precisely identifying the target concept during the denoising process, trained with few-shot learning to require minimal labeled data. Within the identified region, we introduce a training-free Dual Prompts Cross-Attention (DPCA) module to substitute the target concept, ensuring minimal disruption to surrounding content. We evaluate our method on concept localization precision and replacement efficiency. Experimental results demonstrate that our method achieves superior precision in localizing target concepts and performs coherent concept replacement with minimal impact on non-target areas, outperforming existing approaches. The code is available at \url{https://github.com/zhang-lingyun/ConceptReplacer}
\end{abstract}

%% file: sec/1_intro.tex
\section{Introduction}


Powerful new AI models\cite{balaji2022ediffi,saharia2022photorealistic,yu2022scaling,chang2023muse,gafni2022make,xu2023versatile,ding2022cogview2,lu2023tf} are transforming digital image creation. Notably, DALL-E~\cite{ramesh2021zero,ramesh2022hierarchical}, Stable Diffusion~\cite{rombach2022high}, and Midjourney~\cite{midjourney} have met commercial-grade product standards, opening up opportunities for a wide range of user-oriented applications. These models can generate diverse, stunning images from simple text descriptions, redefining how we approach digital art, content creation, and design. However, there’s a major challenge: these models sometimes produce sensitive or inappropriate content. This issue stems from the massive unfiltered datasets they learn from, which inevitably contain inappropriate materials. Since publicly available web-scraped data\cite{changpinyo2021conceptual, schuhmann2022laion} often lack stringent quality control, particularly in terms of bias and safety.


Efficient methods that allow large-scale text-to-image models to selectively remove specific concepts are emerging as a promising avenue. Current approaches to addressing unsafe content generation can be broadly categorized into the following three main strategies: (1) \textit{Dataset-level preprocessing}, as in Stable Diffusion 2.0\cite{stable_diffusion_2}, involves using classifiers to pre-filter images containing sexually explicit content in large datasets like LAION\cite{schuhmann2022laion}. However, this process incurs substantial computational costs, requiring approximately 150,000 GPU hours over the 5-billion-image LAION dataset. Despite these efforts, sexually explicit content may still emerge in model outputs. (2) \textit{Post-generation solutions}, such as the NSFW filter\cite{rando2022red} in Stable Diffusion, employ classification models to detect and block inappropriate content after generation. While straightforward to implement, these approaches often result in poor user experience by replacing entire images with meaningless placeholders when unsafe content is detected, regardless of the extent or location of the problematic content. (3) \textit{Generation-time guidance methods}, including SLD\cite{schramowski2023safe} and ESD-u\cite{gandikota2023erasing}, represent a more dynamic approach by incorporating noise-prediction guidance during the inference or training phase. These methods aim to suppress unsafe content generation through real-time interventions in the diffusion process. However, their effectiveness comes at a cost: the guidance mechanisms typically affect broad regions of the generated image, often modifying unintended areas and compromising the model's ability to generate high-quality, detailed outputs.



To sum up, preventing unsafe content generation in large-scale diffusion models is still a major unresolved challenge. Current methods have key limitations: post-generation filtering hurts user experience, dataset filtering is resource-intensive yet ineffective, and generation-time guidance reduces image quality.
Recent studies\cite{khani2023slime, wang2023diffusion} on segmentation have revealed an encouraging insight: stable diffusion models, with their attention mechanisms, possess an inherent capability to detect and localize objects. This finding suggests that these models might be capable of precisely identifying problematic content without compromising the overall image generation process. However, to our knowledge, there is no existing method that can both precisely locate and replace problematic content while preserving the intended meaning and visual quality of other areas. Motivated by these challenges and opportunities, we introduce Concept Replacer, a novel framework for precise concept replacement in diffusion models. Our approach is built on two key insights: First, given the diverse nature of unsafe content, we design our framework to precisely locate unsafe areas based on just a few examples, leveraging the model's inherent object detection capabilities. Second, we ensure the replacement process is customizable to accommodate different safety requirements and content preferences, allowing for flexible and context-aware content modification. This framework addresses the limitations of existing methods while maintaining generation quality and semantic coherence.

Our Concept Replacer consists of three key components: (1) a concept localizer, built upon a pretrained diffusion model through efficient fine-tuning, which precisely identifies concept locations during the generation process; (2) a Dual Prompts Cross-Attention module that leverages two distinct prompts to guide the replacement of targeted concepts; and (3) an integrated denoising process that combines localization and replacement capabilities for harmonious concept substitution. During the diffusion model's denoising process, our concept localizer, trained using few-shot learning, identifies the location of the target concept in the latent space. Then, our dual prompts cross-attention module processes the original input prompt and the replacement prompt simultaneously. The replacement prompt specifically guides the processing of image features in the target concept area. Importantly, our method maintains consistency in both style and content between the replaced region and the surrounding areas, resulting in a seamlessly integrated final image where the replaced content naturally blends with the original context. Our method outperforms existing methods on accuracy of concept replacing. Furthermore, it is consistent with the output of the original model in the non-correlated regions.


Our primary contributions are as follows:
\begin{itemize}

\item We introduce a few-shot trained concept localizer specifically designed for real-time concept identification during the denoising process, offering efficient and accurate concept detection with minimal training requirements.

\item We introduce an innovative Dual Prompts Cross-Attention module that leverages precise concept localization to enable targeted concept replacement while preserving the surrounding image context.

\item We demonstrate the superiority of our approach through comprehensive quantitative and qualitative evaluations, establishing new benchmarks in both localization accuracy and replacement effectiveness.
\end{itemize}

%% file: sec/2_ralated_work.tex
\section{Related Work}

\subsection{Target Localization in Diffusion Models}
Precisely localizing concepts within diffusion models is crucial for effective concept manipulation. Large-scale pre-trained text-to-image models~\cite{kawar2023imagic,yu2022scaling, saharia2022photorealistic,ramesh2021zero} have enabled advances in image segmentation. DiffSegmenter~\cite{wang2023diffusion} utilizes self-attention and cross-attention in U-Net~\cite{ronneberger2015u} to perform segmentation in a training-free manner. SLiMe~\cite{khani2023slime} trains word embeddings, using few shot learning to achieve part segmentation of target concepts. DIFFEDIT~\cite{couairon2022diffedit} and Watch Your Steps~\cite{mirzaei2025watch} obtain masks for target concepts by predicting differences in noise under different prompts conditioning, with Watch Your Steps leveraging InstructPix2Pix~\cite{brooks2023instructpix2pix} to derive varying noise predictions. All those methods are aimed at localizing objects in real images. 
Inspired by these methods, we explore the concept of localization in the image generation process to achieve precise concept replacement. 

Some works~\cite{wu2023harnessing,feng2022training, chefer2023attend} enhance the controllability of text-to-image diffusion models with an attention mechanism. Another line of work employs object localization with attention to guide the image editing process. Prompt2Prompt~\cite{hertz2022prompt} utilizes layers of cross-attention to manage attributes in the image, requiring both the source and the target commands to share an identical structure. PnP~\cite{tumanyan2023plug} explores the use of attention
and feature injection to improve image-to-image translation. LPM~\cite{patashnik2023localizing} employs self-attention and cross-attention to generate images with variations in the shape of a specific object. FoI~\cite{guo2024focus} extract masks from cross-condition attention in a pretrained IP2P model to execute text-guided real image editing. In contrast to those methods, our approach concentrates on identifying concepts during the image generation phase and replacing them with another concept.


\subsection{Concept Removal in Diffusion Models}
The removal of specific concepts from diffusion models is a critical issue, as large-scale diffusion models can generate undesirable and unsafe content. Currently, there are three main approaches to restricting the generation of images containing target concepts: 
dataset-level preprocessing, post-generation solutions and generation-time guidance methods.

Dataset-level preprocessing filters out unsafe content from the training dataset. This approach~\cite{stable_diffusion_2} normally costs a lot of labor, as it requires filtering large amounts of data and retraining the whole model on the filtered dataset. Post-generation solutions classify the generated images during inference. If the generated image is classified as containing an unsafe concept, it is replaced with a predefined meaningless black image. It relies heavily on the accuracy of the classifier, and in practice, it is challenging to get an accurate Classifier also returning a meaningless image is not user-friendly.

Generation-time guidance methods~\cite{kumari2023ablating, schramowski2023safe,gandikota2023erasing,heng2024selective} can be applied during the inference process or by fine-tuning the model. SLD~\cite{schramowski2023safe} applies positive guidance during inference, introducing a prompt-defined safety direction, and guiding image generation. Ablating concepts~\cite{schramowski2023safe} and Selective Amnesia~\cite{heng2024selective} modify the model's weights to shift the image generation distribution from a target concept to a different user-defined concept. ESD~\cite{gandikota2023erasing} fine-tunes the model to remove a target concept, learning a noise prediction influenced by prompt-defined safety direction, which has the advantages of being fast to implement and difficult to bypass. Mace~\cite{lu2024mace} focuses on tuning the prompts-related projection matrices with LoRA~\cite{hu2021lora} in cross attention layers with a closed-form solution. UCE~\cite{gandikota2024unified} also edits the cross-attention weights without training using a closed-form solution to manipulate concepts in diffusion models. Forget-Me-Not~\cite{zhang2024forget} attempts to address the aforementioned inconsistency issues by incorporating an attention re-steering loss to guide the model's generation away from undesired concepts.
However, those concept removal methods are based on global guidance, 
which affects unrelated areas of the generated image and results in an output that could diverge from the original model. Moreover,
modifying the diffusion U-Net through fine-tuning continues to create discrepancies with the original model.

Different from the aforementioned approaches, we present a method called Concept Replace, which does not require fine-tuning of the original diffusion U-Net. Instead, it employs a concept localizer to locate concepts during the denoising process. This enables us to substitute the targeted concept within its specific area with our training-free Dual Prompts Cross Attention while leaving the surrounding non-target areas unchanged.

%% file: sec/3_method.tex
\section{Method}

\begin{figure*}[h]
  \centering
  \includegraphics[width= 0.9 \linewidth]{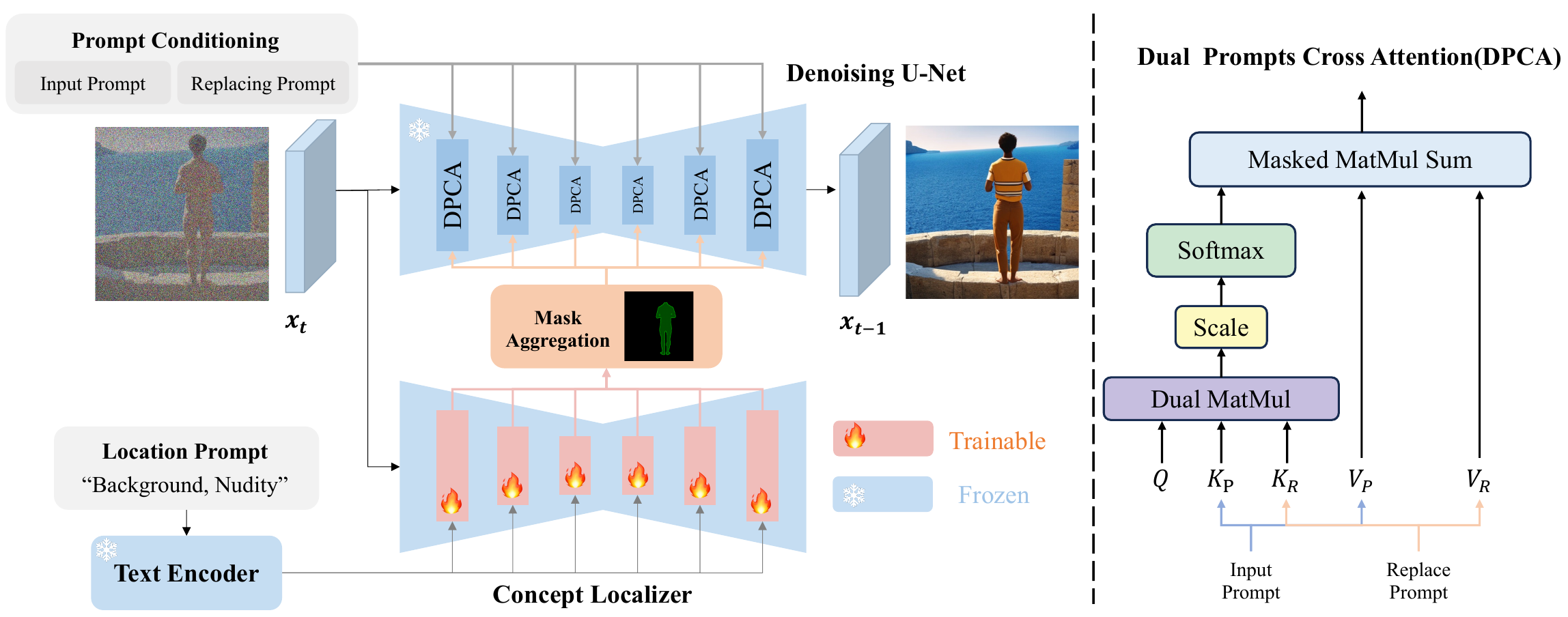}
    \vspace{-0.1in}
  \caption{Framework of Our Method. Our approach is designed to replace a specified target concept during image generation within diffusion models. First, our few-shot trained concept localizer identifies the target concept’s precise location. Next, in the Dual Prompts Cross-Attention module, the target concept is replaced, guided by both the input and replacing prompts. The replacing prompt serves as conditioning specifically for the target concept’s localized area within the image features. Our Dual Prompts Cross-Attention module is training-free, seamlessly replacing the target concept during the denoising phase of diffusion models without affecting non-target regions.}
  \label{fig:figure1}
    \vspace{-0.1in}
\end{figure*}

Diffusion models~\cite{ho2020denoising} are powerful generative models and designed to learn data distribution \( p(x) \) by gradual denoising a Gaussian distribution. Starting from sampled Gaussian noise, the diffusion model gradually denoises for \( T \) time steps to generate the final image:
\begin{equation}
    p_{\theta}(x_{T:0}) = p(x_T) \prod_{t=T}^{1} p_{\theta}(x_{t-1} \mid x_t)
    \label{eq1}
\end{equation}
where \( p(x_T) \)  corresponds to the initial Gaussian noise and \( p(x_0) \) corresponds to the final generated image. Our goal is to remove a target concept  \( c \) during the denoising process of the image \( p(x_0) \) being generated. 

The motivation is to replace target concepts in a diffusion model during the denoising process based on accurate localization of the target concept, avoiding impact on non-target regions of the generated image. Previous methods for removing concepts from diffusion models often affect the whole output, as fine-tuning the diffusion model or introducing guidance during the inference tends to influence the entire output. To solve this problem, we present a concept replacing method based on precise localization. By constructing a dedicated concept localizer to locate the target concept, we replace the concept within the localized region with our proposed Dual Prompts Cross Attention module. 

As shown in Figure \ref{fig:figure1}, our method consists of two main components: a dedicated concept localizer, which is used to localize the target concept during the denoising process. And a novel Dual Prompts Cross Attention module, which allows the original prompt and the replacing prompt to condition the image features based on the localization information, enabling concept replacement in the target area. 

\subsection{Concept Localizer}


\begin{figure}[h]
  \centering
  \includegraphics[width= 1.0 \linewidth]{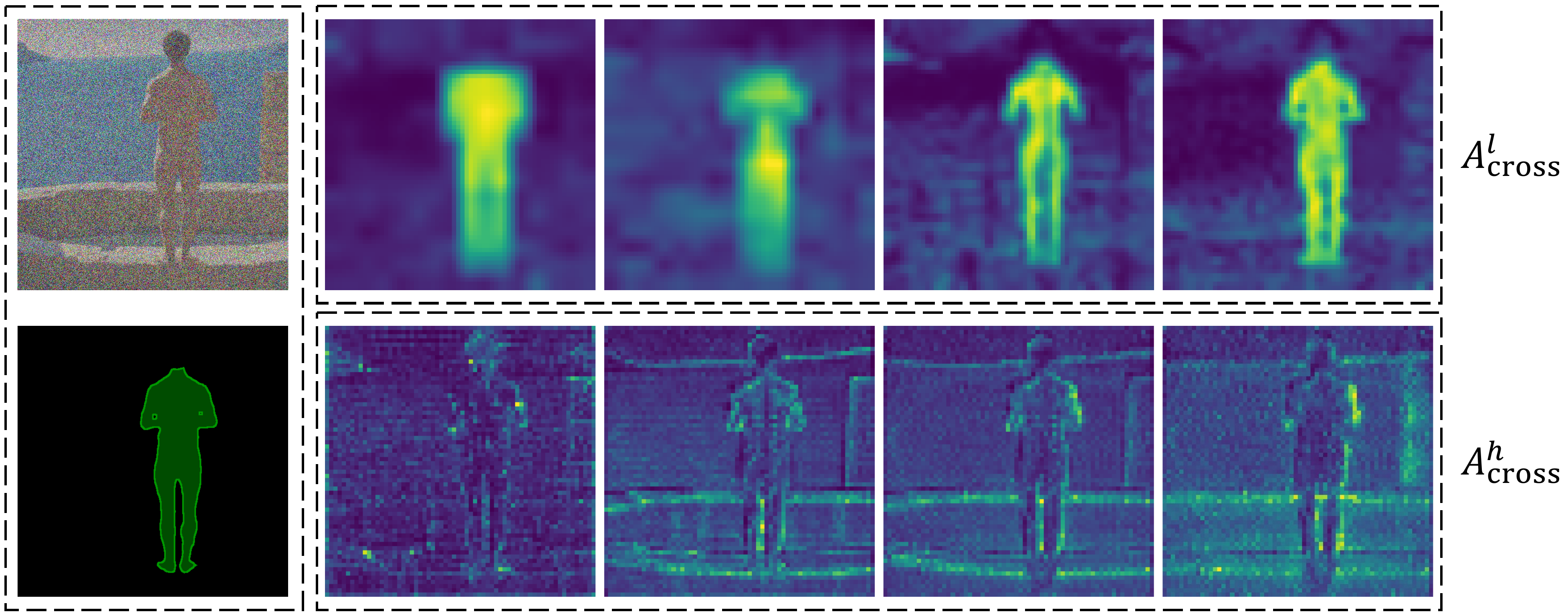}
    \vspace{-0.2in}
  \caption{Visualization of Cross-Attention Maps at Different Spatial Resolutions at Various Levels for the Target Concept. Cross-attention maps at varying spatial resolutions capture distinct types of information for the target concept. Maps \( A^{l}_{cross} \) with smaller spatial dimensions primarily capture low-frequency semantic information, while maps \( A^{h}_{cross} \) with larger spatial dimensions retain high-frequency, fine-grained details. }
  \label{fig:figure2}
  \vspace{-0.1in}
\end{figure}

\begin{figure}
  \centering
  \includegraphics[width= 0.95 \linewidth]{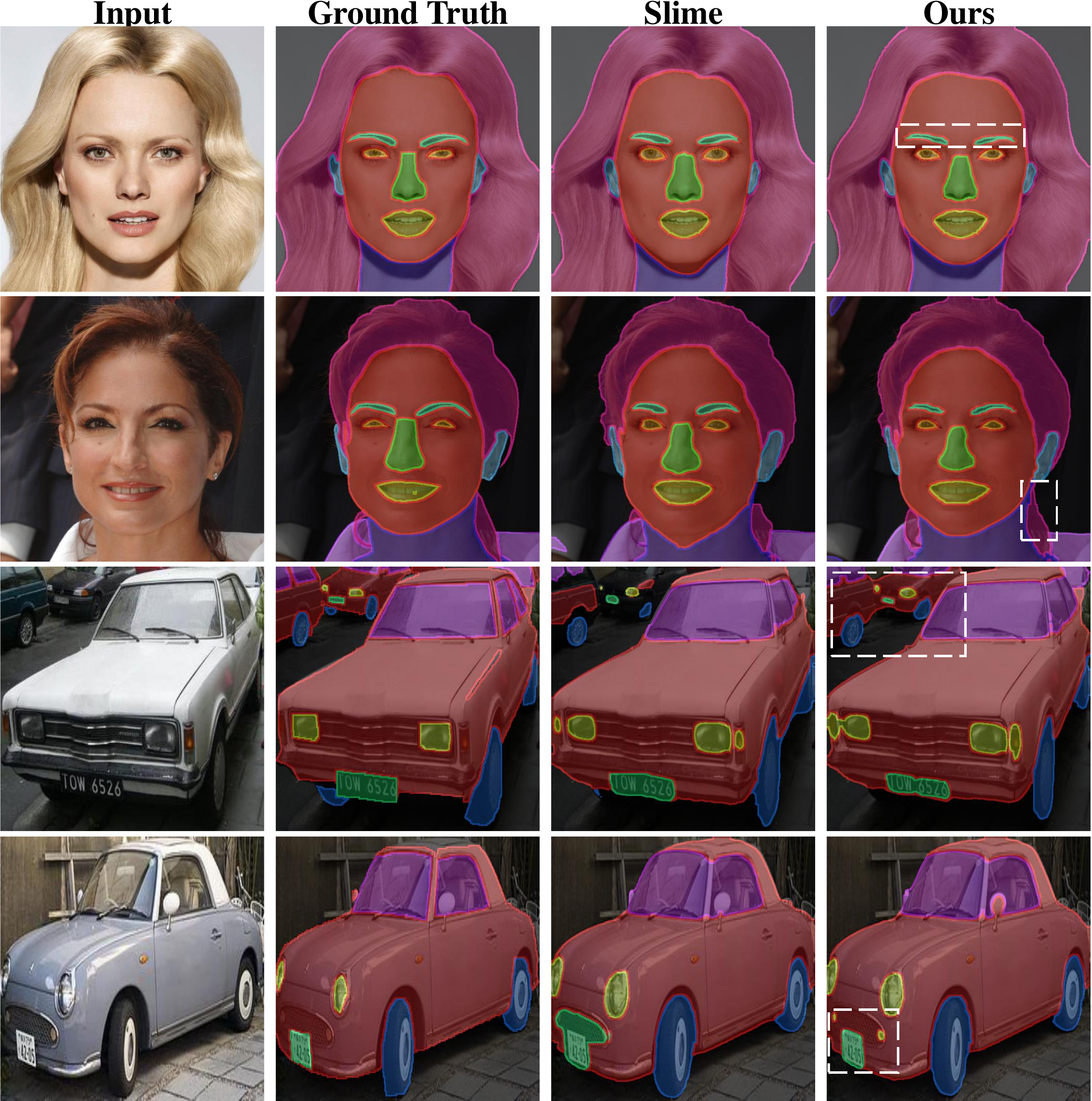}
  \vspace{-0.1in}
  \caption{Segmentation Results on CelebAMask-HQ and Pascal-Car. Our concept localizer is compared with SLiMe for real-image segmentation, showing that our method achieves superior detail accuracy. }
  \label{fig:figure3}
    \vspace{-0.2in}
\end{figure}

To ensure that non-target regions remain unaffected during concept removal, we designed a dedicated concept localizer to localize the target concept during the denoising process of image generation. To avoid the labor and labeling data required to train a locator from scratch, we make full use of the knowledge from the pre-trained U-Net in diffusion models. 


As shown in Figure \ref{fig:figure1}, the concept localizer takes a location prompt and the image embedding as input and outputs the location mask corresponding to the target concept represented by the location prompt. To fully utilize the knowledge of the pre-trained U-Net of diffusion models, our concept localizer shares the same structure as the original U-Net and we fine-tune projection matrices \( W_k \) and \( W_v \)  in the self-attention layers and cross-attention layers. Then attention scores are extracted from these self-attention and cross-attention and further fused to get the final mask as the location of the target concept.

For each attention layer, given the query \( Q \) and the key \( K \), we extract its attention score as:
\begin{equation}
    \text{A} = \text{Softmax}(\frac{Q K^T}{\sqrt{d}})
    \label{eq2}
\end{equation}
where \( d \) is the output dimension of key and query features.

The U-Net of diffusion models has multiple cross attention layers and self attention layers. First, We average all the attention scores that have been resized to the same size from self attention layers:
\begin{equation}
A_{self} = \sum_{n \in N} (\text{Resize}(\{A^{n}_{self}\}))
\label{eq3}
\end{equation}
where \( N \) represents the number of self-attention layers.
In our experiments, we discovered that the lower-resolution cross-attention scores possess better spatial semantic disentanglement, while the higher-resolution cross attention scores exhibit weaker spatial semantic disentanglement as shown in Figure~\ref{fig:figure2}. Thus, directly averaging them can decrease the accuracy of the final mask. However, higher cross attention scores contain more high-frequency image details.
For cross attention layers,  we separate the cross attention scores according to the size of the resolution and separately average them to get \( A^{l}_{cross} \) and \( A^{h}_{cross} \) as the same as Eq.\ref{eq3} but with different cross attention layers.
\( l \) stands for the lower resolution cross-attention layers, and \( h \) refers to the high resolution cross-attention layers.  In our experiments, we empirically divided the cross attention layer whose spatial dimension is less than \(32 \times 32 \) into low resolution cross-attention layers and others as high resolution cross attention layers.
We refine the higher cross attention scores using the lower cross attention scores, which helps to ensure that the final mask retains accurate details while maintaining semantic accuracy.
\begin{figure}
  \centering
  \includegraphics[width= 1. \linewidth]{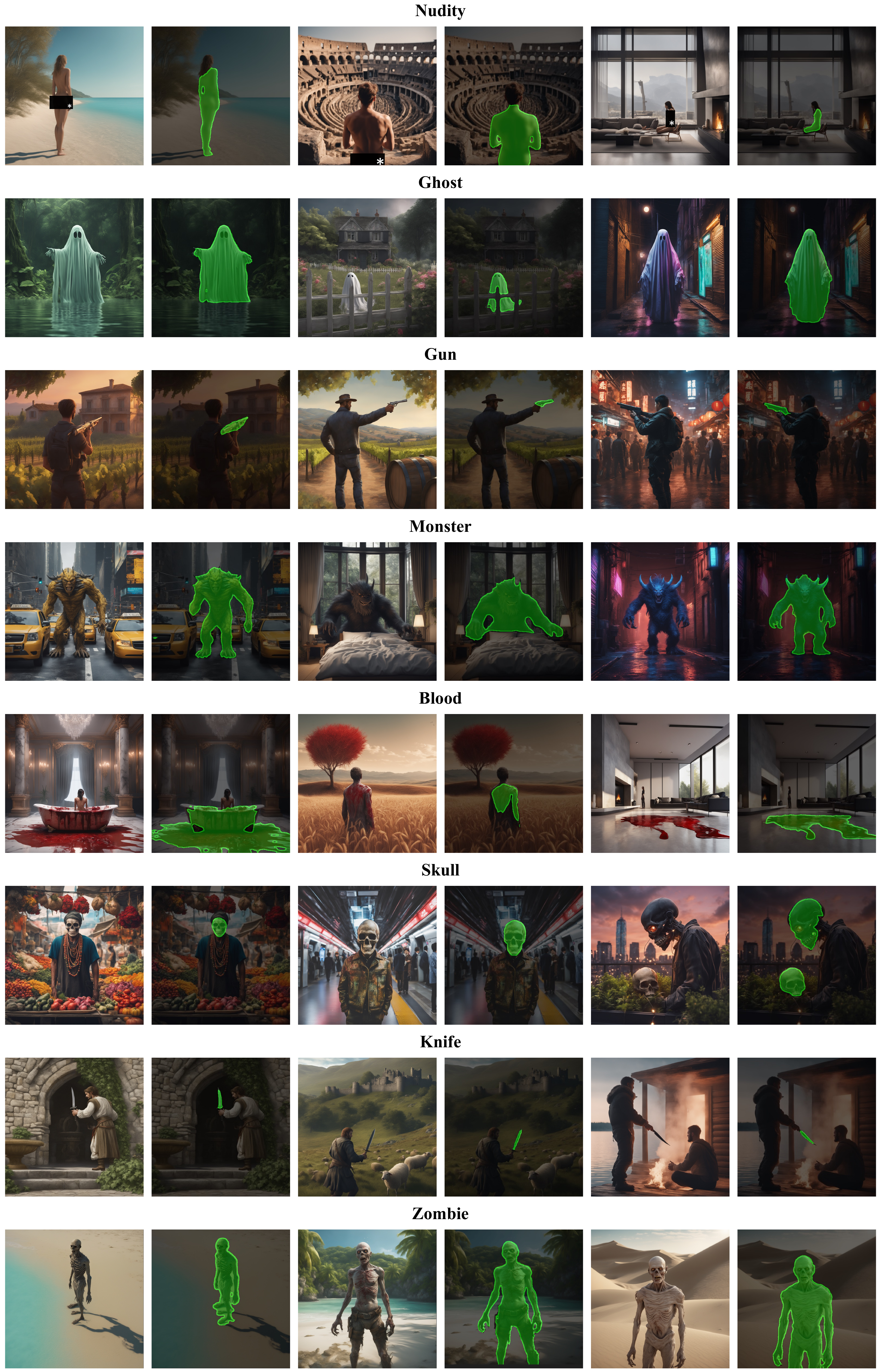}
  \caption{Concept Localization Results with the Proposed Concept Localizer. Our method effectively pinpoints target concepts during image generation, accurately identifying objects across varying sizes.}
  \label{fig:figure4}
\end{figure}
\begin{figure*}
  \centering
  \includegraphics[width= 0.95 \linewidth]{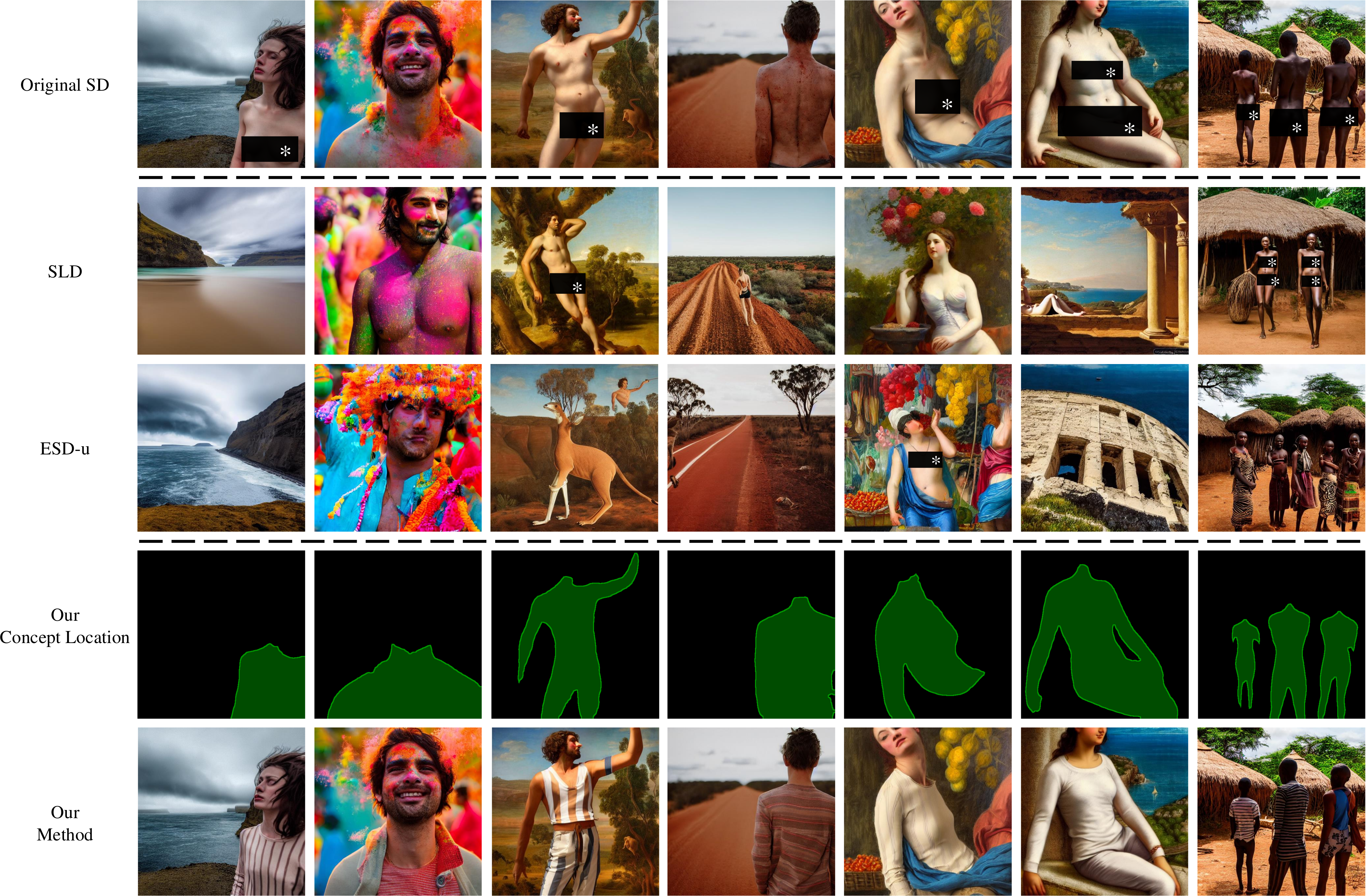}
    \vspace{-0.1in}
  \caption{Nudity Concept Replacement Results. Unlike other methods, our approach identifies the target concept during image generation, allowing precise replacement while preserving the consistency of non-target areas with the original model. }
  \label{fig:figure5}
    \vspace{-0.1in}
\end{figure*}
The final cross attention score is defined as follows:
\begin{equation}
    A_{cross} = A^{l}_{cross} + A^{l}_{cross} \cdot A^{h}_{cross}
    \label{eq:placeholder}
\end{equation}
Previous methods like SLiMe~\cite{khani2023slime} and DiffSegmenter~\cite{wang2023diffusion} have demonstrated the importance of self-attention in segmentation tasks. Following those works, we further combine self-attention scores and cross-attention scores to obtain the final attention score. This further leverages self-attention to capture spatial relationships and cross-attention to refine the semantic alignment between the location prompt and the image embedding. By integrating both types of attention scores, we enhance the precision of the target concept localization.
\begin{equation}
    M = vec(A_{cross}) \ast A_{self}
    \label{eq:example}
\end{equation}
where \( vec \) denotes the vectorization of matrix \( A_{cross} \), which stacks all rows of \( A_{cross} \) into a single row vector.
During the training of concept localizer with few-shot examples, we apply cross-entropy loss to the cross-attention score:
\begin{equation}
    \mathcal{L}_{CE} = CE(M, M')
    \label{eq:cross_entropy}
\end{equation}
where \( M' \) represents the segmentation label and CE refers to cross-entropy.
Additionally, for the final mask \( M \), which combines both self-attention and cross-attention scores, we utilize mean squared error (MSE) loss to refine the accuracy of the mask. 
\begin{equation}
    \mathcal{L}_{MSE} = \left\| \mathbf{M} - \mathbf{M}'_k \right\|_2^2
    \label{eq:placeholder}
\end{equation}

\subsection{Dual prompts cross attention}
The text-to-image diffusion models enable conditioning on the prompt by augmenting U-Net with cross attention mechanism. Recall that the original cross-attention~\cite{vaswani2017attention} of diffusion models is defined as:

\begin{equation}
    Z =  \text{Softmax} \left( \frac{Q K^T}{\sqrt{d}} \right) V
    \label{eq:cross_attention}
\end{equation}
where key \( K \) and value \( V \) are derived from the text embedding and query \( Q \) is derived from the image embedding. 

To replace the target concept within the location, we propose a Dual Prompts Cross Attention module that additionally takes a concept location mask and a replacing prompt. The replacing prompt specifies the concept that will replace the target concept in the identified areas.

With the concept location mask \( M \)  and a replace prompt, our Dual Prompts Cross Attention is defined as follows:
\begin{equation}
\begin{aligned}
Z &= \text{Softmax}\left(\frac{Q  \cdot K_R^T}{\sqrt{d}}\right)V_R \cdot M \\
&\quad + \text{Softmax}\left(\frac{Q \cdot K_P^T}{\sqrt{d}}\right)V_P \cdot (1 - M)
\end{aligned}
\end{equation}
where \( K_P \) and \( V_P \) denote the key and value derived from the input prompt, and \( K_R \), \( V_R \) as the key and value for the replacing prompt. Our Dual Prompts Cross Attention module allows the concept to be replaced according to the mask without requiring any additional training, thereby generating the image where the target concept has been replaced. It ensures a seamless replacement process by using different prompt conditioning for different areas of the image embedding.

During the inference phase, our concept localizer is activated in 2-3 time steps of the initial stage of denoising to detect the location of the target concept. Once detected, our Dual Prompts Cross-Attention module engages to replace the target concept. Otherwise, the process proceeds identically to the original Stable Diffusion model. This further allows for targeted concept replacement while preserving the model’s original image generation capabilities.

%% file: sec/4_experiments.tex
\begin{table*}
  \centering
  \begin{tabular}{lccccccccccc}
    \toprule
     & Cloth & Eyebrow & Ear & Eye & Hair & Mouth & Neck & Nose & Face & Background & Average \\
     \hline ReGAN & 15.5 & \textbf{68.2} & 37.3 & \textbf{75.4} & 84.0 & \textbf{86.5} & 80.3 & \textbf{84.6} & 90.0 & 84.7 & 69.9 \\
     SegDDPM & 61.6 & 67.5 & \textbf{71.3} & 73.5 & 86.1 & 83.5 & 79.2 & 81.9 & 89.2 & 86.5 & 78.0 \\
     SLiMe & 63.1  & 62.0  & 64.2  & 65.5  & 85.3  & 82.1  & 79.4 & 79.1  & 88.8  & 87.1 & 75.7  \\
     \textbf{Ours} & \textbf{67.1} & 63.7 & 65.7 & 72.6 & \textbf{86.4} & 83.0 & \textbf{82.5} & 81.0 & \textbf{90.0} & \textbf{87.9} & \textbf{78.1} \\
     \hline ReGAN & - & - & - & 57.8 & - & 71.1 & - & 76.0 & - & - & - \\
     SegGPT* & 24 & \textbf{48.8} & 32.3 & 51.7 & \textbf{82.7} & 66.7 & 77.3 & 73.6 & 85.7 & 28.0 & 57.1  \\
     SegDDPM & 28.9 & 46.6 & 57.3 & 61.5 & 72.3 & 44.0 & 66.6 & 69.4 & 77.5 & 76.6 & 60.1 \\
     SLiMe & \textbf{52.6}  & 44.2  & 57.1  & 61.3  & 80.9  & 74.8  & 78.9 & 77.5  & \textbf{86.8} & 81.6  & 69.6  \\
     \textbf{Ours} &  38.9 & 42.1 & \textbf{65.0} & \textbf{66.1} & 81.3 & \textbf{79.9} & \textbf{79.0} & \textbf{81.7} & 85.8 & \textbf{81.7} & \textbf{70.2} \\
    \bottomrule
  \end{tabular}
  \caption{CelebA-Mask-HQ Segmentation Results. The first three rows display results with 10 training samples, and the following five rows show results with 1 training sample. Methods marked with * indicate supervised approaches. Overall, our method consistently achieves superior or comparable performance across most instances and on average.}
  \label{tab:1}
\end{table*}

\begin{figure*}[htbp]
  \centering
  \includegraphics[width= 1. \linewidth]{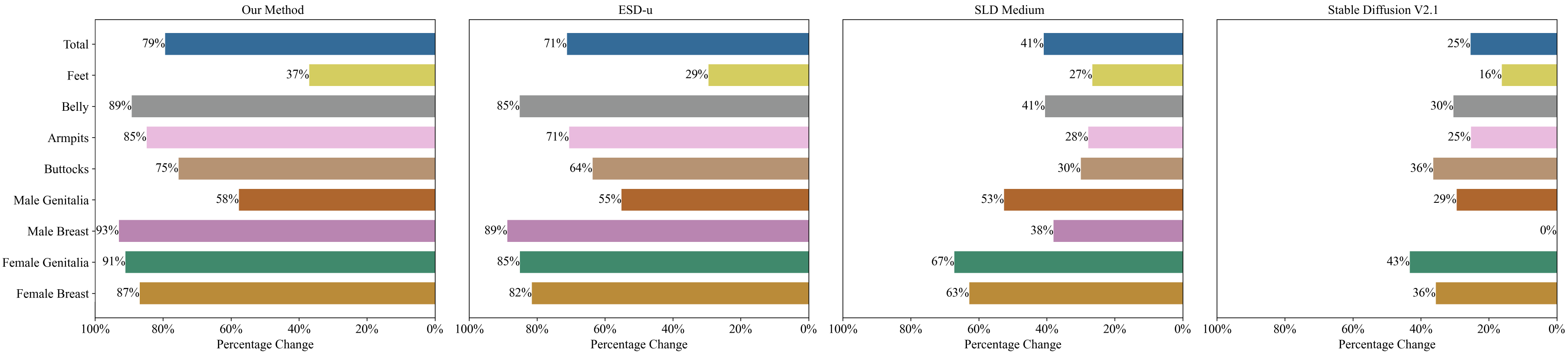}
    \vspace{-0.2in}
  \caption{Results of Nudity Concept Removal. We present percentage reductions in nudity content relative to original Stable Diffusion v1.4 on the I2P prompts dataset. Higher percentages represent more effective removal. Our approach effectively reduces nudity-related content in Stable Diffusion, outperforming other methods.}
  \label{fig:figure6}
    \vspace{-0.1in}
\end{figure*}

\begin{figure*}
  \centering
  \includegraphics[width= 1.0 \linewidth]{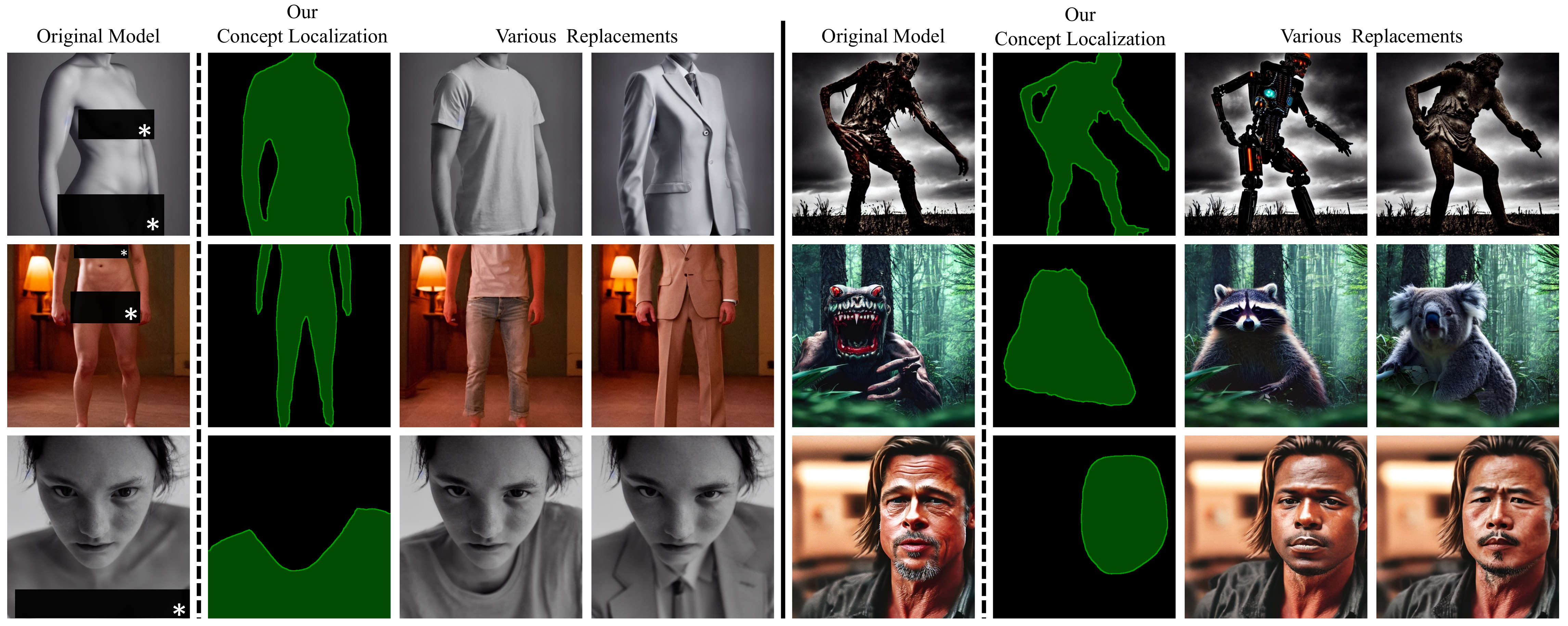}
  \caption{Concept Replacement with Various Replacing Prompts. Our method accurately identifies the specified concept and seamlessly substitutes it with a new concept, as defined by the replacing prompt, during the image generation process. Zoom in for details. More examples are in Supplementary Material.}
  \label{fig:figure7}
\end{figure*}

\begin{table*}
  \centering
  \begin{tabular}{lccccccc}
    \toprule
     & Body & Light & Plate & Wheel & Window & Background & Average \\
     \hline $\mathrm{CNN}^{\star}$ & 73.4 & 42.2 & 41.7 & 66.3 & 61.0 & 67.4 & 58.7 \\
     CNN+CRF* & 75.4 & 36.1 & 35.8 & 64.3 & 61.8 & 68.7 & 57 \\
     \hline ReGAN & 75.5 & 29.3 & 17.8 & 57.2 & 62.4 & 70.7 & 52.15 \\
     SLiMe & 81.5 & 56.8  & \textbf{54.8} & 68.3 & 70.3 & 78.4 & 68.3 \\
     \textbf{Ours} & \textbf{82.26} & \textbf{59.41} & 52.55 & \textbf{69.70} & \textbf{72.21} & \textbf{79.59} & \textbf{69.29} \\
     \hline SegGPT* & 62.7 & 18.5 & 25.8 & 65.8 & 69.5 & 77.7 & 53.3 \\
     SLiMe & 79.6 & 37.5 & \textbf{46.5} & 65.0 & 65.6 & 75.7 & 61.6 \\
     \textbf{Ours} & \textbf{79.6} & \textbf{51.8} & 43.5 & \textbf{66.5} & \textbf{65.6} & \textbf{79.0} & \textbf{64.3}\\
    \bottomrule
  \end{tabular}
  \caption{Pascal-Car Segmentation Results. The first two rows display results from supervised methods, followed by the next three rows showing the performance with 10-sample training, and the final three rows illustrating the 1-sample training setting. Our concept localizer consistently outperforms other methods across most classes and on average.}
  \label{tab:2}
\end{table*}

\section{Experiments}

To validate our method, we conducted both quantitative and qualitative analyses to evaluate the accuracy of the localization of our concept localizer and the effectiveness of the concept replacement.

\subsection{Concept Localization Experiments}
This section focuses on assessing the precision of our concept localizer. 
We evaluate localization accuracy through image segmentation, as it effectively demonstrates the precision of our concept localizer, even though our method is primarily designed for concept localization during diffusion model generation.

\textbf{Dataset.} We validate our localization accuracy on the CelebAMask-HQ dataset~\cite{lee2020maskgan} and Pascal-Car dataset~\cite{chen2014detect}. Following SLiMe~\cite{khani2023slime}, we train the model with both 1-shot and 10-shot settings. 

\textbf{Comparison Methods.} We compare our method with the state-of-the-art approaches, including ReGAN~\cite{tritrong2021repurposing}, SegDDPM~\cite{baranchuk2021label}, SegGPT~\cite{wang2023seggpt}, and SLiMe~\cite{khani2023slime}. ReGAN and SegDDPM necessitate an initial model pre-training phase on the dataset before tackling segmentation tasks. ReGAN relies on a pre-trained GAN model~\cite{goodfellow2014generative}, while SegDDPM utilizes pre-training on a DDPM~\cite{ho2020denoising}. In both cases, pre-training is executed on specific datasets. Conversely, SegGPT employs several segmentation datasets for supervised training, demanding a substantial volume of training data. SLiMe employs few-shot learning on a pre-trained stable diffusion model, optimizing word embeddings for segmentation. Similarly, our localization module also employs few-shot learning, leveraging the understanding of concepts from stable diffusion without relying on extensive labeled data.

\textbf{Evaluation Metrics.} To evaluate localization accuracy, we compute the mean intersection over union (mIOU) for each of the categories on the CelebAMask-HQ test set and Pascal-Car test set, and also calculate the average mIOU across all categories to measure the overall accuracy.

Table~\ref{tab:1} presents the quantitative experimental results under the 10-shot and 1-shot training settings on the CelebA-Mask-HQ. It is shown our method outperforms ReGAn, SegDDPM and SLiMe in the majority class and on average in both the 1-shot and 10-shot settings. Likewise, Table~\ref{tab:2} displays our results for the car datasets and Figure~\ref{fig:figure3} qualitatively shows our results on both datasets. SegGPT is trained in a supervised manner on large segmentation datasets, while the other two methods require pre-training on specific categories of data. Both SLiMe and our method rely on few-shot training with a trained stable diffusion model. Our approach attains more precise image segmentation by fully fine-tuning self attention layers and cross attention layers of stable diffusion. Figure~\ref{fig:figure4} shows the location of multiple concepts of varying sizes. It demonstrates our method's capability to identify concepts of different magnitudes.


\subsection{Content Replacement Experiments}
In this section, we validate the effectiveness of our method to replace target concepts.

\textbf{Dataset.} To quantitatively evaluate the effect of concept replacement of our method, we generate images on the I2P prompt dataset~\cite{schramowski2023safe} with stable diffusion. The I2P dataset comprises 4,073 prompts with a strong likelihood of producing unsafe material. We use this dataset to generate images, with a focus on removing nudity as the target concept, in order to evaluate the replacement efficiency. To further evaluate the impact of our method on the model’s ability to generate standard content, we evaluate image quality using the COCO 30k prompts dataset~\cite{lin2014microsoft} which is a well
curated dataset without nudity.

\begin{table} \small
  \centering
  \begin{tabular}{lcc}
    \toprule
    Method & FID-30k ${\downarrow}$ & CLIP ${\downarrow}$  \\
    \midrule
    REAL & - & 30.41 \\
    SD & 14.50 & 31.32 \\
    SLD-Medium  & 16.90 & 30.46 \\
    ESD-u & 14.16  & 30.45\\
    Ours & 15.15 & 30.67 \\
 \bottomrule
  \end{tabular}
  \caption{Image Fidelity and Text Alignment Results on COCO 30K Dataset. All methods produce similar results, indicating that the impact on image quality and text alignment in the COCO 30K dataset is minimal.}
  \label{tab:3}
\end{table}

\begin{table*}
  \centering
  \begin{tabular}{lccccccccccc}
    \toprule
     & Cloth & Eyebrow & Ear & Eye & Hair & Mouth & Neck & Nose & Face & Background & Average \\
     \hline 
     Ours-L &  64.5 & 63.8 & 65.6 & 72.0 & 86.2 & 83.3 & 81.0 & 82.0 & 90.1 & 86.6 &  77.5  \\
     Ours-H & 66.8 &  63.4     & 64.4   & 73.2  & 85.9 &  83.0 &   82.1 & 81.7 & 90.1 & 87.7 &  77.8 \\
     Ours-T & 67.1 & 63.7 & 65.7 & 72.6 & 86.4 & 83.0 & 82.5 & 81.0 & 90.0 & 87.9 & 78.1 \\
    \bottomrule
  \end{tabular}
  \caption{Ablation Study on Concept Localizer with Different Configurations. }
  \label{tab:suppl_1}
\end{table*}

\textbf{Comparison Methods.} We compare our method with stable diffusion v2.1~\cite{stable_diffusion_2}, SLD~\cite{schramowski2023safe}, and ESD~\cite{gandikota2023erasing}. Stable diffusion v2.1 trained on filtered datasets that filter out the NSFW images. SLD removes a concept by introducing positive guidance during the image inference process. ESD removes a concept by fine-tuning the entire model.

\textbf{Evaluation Metrics.} We use NudeNet~\cite{nudenet} to detect nudity-related content in the generated images to evaluate the effectiveness of removing the specified concept of nudity. The FID and CLIP~\cite{radford2021learning} scores are used to assess the method’s impact on normal content with image fidelity and text alignment.

Figure~\ref{fig:figure5} shows the results of replacing the nudity concept, demonstrating that our method accurately locates and seamlessly replaces it. Notably, the non-target areas remain consistent with the original Stable Diffusion model, outperforming other methods. Figure~\ref{fig:figure6} presents the quantitative results of removing the nudity concept from the I2P prompt datasets. We generate images using the I2P prompt datasets and employ NudeNet to detect nudity in the generated images. Our method shows a significant reduction in nudity content compared to the original Stable Diffusion v1.4 model. Across all categories identified by NudeNet, our approach consistently outperforms others, achieving a greater reduction in censored images, thereby demonstrating superior effectiveness in removing the nudity concept. Table~\ref{tab:3} presents the results on the COCO dataset, showing that all methods have minimal impact on image quality  and text alignment.
Figure~\ref{fig:figure7} illustrates the replacing of concepts with various replacements. Our method efficiently replacing concepts by employing various prompts, demonstrating the success of our approach.
Collectively, these results validate the effectiveness of our approach in achieving accurate localization and harmonious replacement, reinforcing its potential for targeted concept manipulation in diffusion models.



\begin{figure}[htbp]
  \centering
  \includegraphics[width= 1. \linewidth]{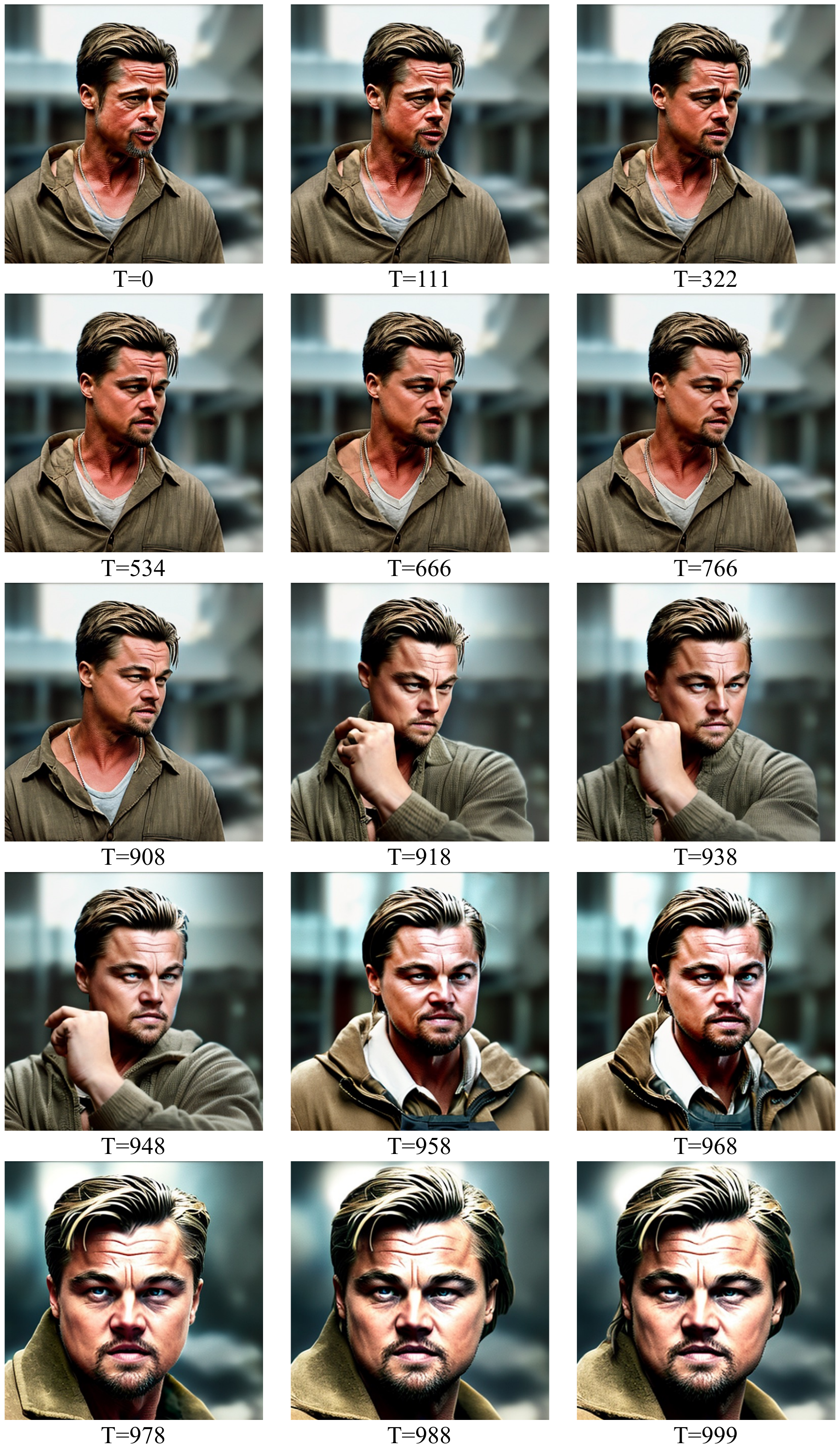}
  \caption{Impact of Replacing Timesteps. Replacing Brad Pitt with Leonardo DiCaprio when \( T < 900 \) results in minimal alteration to the overall structure, while substitution for \( T > 900 \) leads to more significant semantic modifications. This occurs because the early diffusion phase with high \( T \) produces broad semantic content, whereas the later phase with smaller \( T \) focuses on fine details. We selected \( T=666 \) as the optimal moment for substitution to preserve the overall structure and substitution impact effectively.}
  \label{fig:figure8}
\end{figure}

%% file: sec/5_ablation.tex
\section{Ablation Study}

In this section, we present the ablation studies on concept location and replacing, which illustrate the impact of each design.   

\textbf{Ablation on Concept Localizer.} In Table \ref{tab:suppl_1}, we present the ablation study results on the CelebA-Mask-HQ datasets using 10-shot training. "Ours-L" denotes the concept localizer that employs low-resolution cross-attention layers with spatial dimensions under 32. "Ours-H" signifies the concept localizer incorporating both refined low and high-resolution cross-attention layers. "Ours-T" represents our final concept localizer, combining the "Ours-H" setup with average timesteps. 
We utilize the average of \(T=5, 50, 100\) timesteps for real image segmentation. For concept localization during the denoising process, we calculate the average over \(T=666, 726, 766\) timesteps. This choice is guided by the requirement that concept replacement must occur during the early stages of the denoising process, as illustrated in Figure \ref{fig:figure8}.

\textbf{Ablation on Replacing Timestep.} In Figure \ref{fig:figure8}, we demonstrate different timesteps utilized for replacing the concept "Brad Pitt" with "Leonardo DiCaprio." Among 1000 timesteps, we picked specific points for this replacing process. For \( T=0 \), it refers to the initial image generated using the prompt "a photo of Brad Pitt." From \( T=0 \) to \( T=900 \), there is relatively minimal semantic change, whereas for \( T \) exceeding 900, the semantic alteration becomes significant. This suggests that low-frequency semantic information is established early in the denoising process when \( T \) is large, while high-frequency details emerge when \( T \) is small. 
In our experiments, we replaced the concept at  \( T=666 \) to achieve a balance between the replacement effect and the preservation of the overall structure.

%% file: sec/6_conclusion.tex
\section{Conclusion}
In this study, we introduce Concept Replacer, a method for replacing specific concepts in text-to-image diffusion models via precise localization. 
Our method uses a few-shot trained concept localizer to accurately identify target concepts and our training-free Dual Prompts Cross-Attention module replaces the target concept using localization information, ensuring that non-target regions of the generated image remain unaffected.
Our experiments demonstrate that our method excel in concept localization accuracy and replacement quality compared to existing approaches. 
We believe that our method can serve as a crucial tool for generative models, enabling them to effectively remove diverse unwanted concepts without compromising user experience. 
